\documentclass[sigconf]{acmart}
\usepackage{multirow}
\usepackage{tabularx}
\usepackage{threeparttable}
\usepackage{enumitem}
\usepackage{nicefrac}       % compact symbols for 1/2, etc.
\usepackage{microtype}      % microtypography
\usepackage{xcolor}         % colors
\usepackage{multirow}
\usepackage{tabularx}
\usepackage{threeparttable}
\usepackage{enumitem}
\usepackage{algorithm}
\usepackage{setspace}
\usepackage{algorithmicx}
\usepackage{latexsym}
\usepackage{graphicx}
\usepackage{amsmath}
\usepackage{algorithm}
\usepackage{algorithmicx}
\usepackage{algpseudocode}
\usepackage{multirow}
\usepackage{tabularx}
\usepackage{amsfonts}  
\usepackage{threeparttable}
\usepackage{enumitem}
\usepackage{amsmath}
\usepackage{graphicx}
\usepackage{multirow}
\usepackage{threeparttable}
\usepackage{tabularx}
\usepackage{algorithm}
\usepackage{setspace}
\usepackage{hyperref} 
\usepackage{xcolor}
\usepackage{soul}
\usepackage{cleveref}
\usepackage{url}
\usepackage{colortbl}

\definecolor{green1}{RGB}{223,240,214}

\definecolor{yellow1}{RGB}{247,214,127}

\AtBeginDocument{%
  \providecommand\BibTeX{{%
    \normalfont B\kern-0.5em{\scshape i\kern-0.25em b}\kern-0.8em\TeX}}}

\setcopyright{acmcopyright}
\copyrightyear{2024} 
\acmYear{2024} 
\setcopyright{acmlicensed}\acmConference[SIGIR '24]{Proceedings of the 47th International ACM SIGIR Conference on Research and Development in Information Retrieval}{July 14--18, 2024}{Washington, DC, USA}
\acmBooktitle{Proceedings of the 47th International ACM SIGIR Conference on Research and Development in Information Retrieval (SIGIR '24), July 14--18, 2024, Washington, DC, USA}
\acmDOI{10.1145/3626772.3657698}
\acmISBN{979-8-4007-0431-4/24/07}

\begin{document}

% \title{Interventional Rationalization}
% \title{Incorporating Causal Intervention into Selective Rationalization}
\title{Event Grounded Criminal Court View Generation with Cooperative (Large) Language Models}


\author{Linan Yue}
\affiliation{%
  \institution{State Key Laboratory of Cognitive Intelligence, University of Science and Technology of China }
  \city{Hefei}
  \country{China}
}
\email{lnyue@mail.ustc.edu.cn}

\author{Qi Liu}
\authornote{Corresponding author.}
\affiliation{%
  \institution{State Key Laboratory of Cognitive Intelligence, University of Science and Technology of China \& Institute of Artificial Intelligence, Hefei Comprehensive National Science Center}
  \city{Hefei}
  \country{China}
}
\email{qiliuql@ustc.edu.cn}

\author{Lili Zhao}
\affiliation{%
  \institution{State Key Laboratory of Cognitive Intelligence, University of Science and Technology of China }
  \city{Hefei}
  \country{China}
}
\email{liliz@mail.ustc.edu.cn}

\author{Li Wang}
\affiliation{%
  \institution{ByteDance}
  \city{Hangzhou}
  \country{China}
}
\email{wl063@mail.ustc.edu.cn}

\author{Weibo Gao}
\affiliation{%
  \institution{State Key Laboratory of Cognitive Intelligence, University of Science and Technology of China }
  \city{Hefei}
  \country{China}
}
\email{weibogao@mail.ustc.edu.cn}

\author{Yanqing An}
\affiliation{%
  \institution{State Key Laboratory of Cognitive Intelligence, University of Science and Technology of China }
  \city{Hefei}
  \country{China}
}
\email{anyq@mail.ustc.edu.cn}


\begin{abstract}
  % \vspace{-0.5cm}
  With the development of legal intelligence, Criminal Court View Generation has attracted much attention as a crucial task of legal intelligence, which aims to generate concise and coherent texts that summarize case facts and provide explanations for verdicts. Existing researches explore the key information in case facts to yield the court views. Most of them employ a coarse-grained approach that partitions the facts into broad segments (e.g., verdict-related sentences) to make predictions. However, this approach fails to capture the complex details present in the case facts, such as various criminal elements and legal events. To this end, in this paper, we propose an Event Grounded Generation (EGG) method for criminal court view generation with cooperative (Large) Language Models, which introduces the fine-grained event information into the generation. Specifically, we first design a LLMs-based extraction method that can extract events in case facts without massive annotated events. Then, we incorporate the extracted events into court view generation by merging case facts and events. Besides, considering the computational burden posed by the use of LLMs in the extraction phase of EGG, we propose a LLMs-free EGG method that can eliminate the requirement for event extraction using LLMs in the inference phase. Extensive experimental results on a real-world dataset clearly validate the effectiveness of our proposed method. Code is available at \url{https://github.com/yuelinan/Codes-of-EGG}.
\end{abstract}





\maketitle

%   \centering
%   \setlength{\belowcaptionskip}{0.cm}
%  \setlength{\abovecaptionskip}{-0.2cm}
%   \includegraphics[width = 14.05cm]{figure/case123.pdf}
%  \caption{
%   % Examples of selective rationalization on the charge prediction. Annotated rationales are \underline{underlined}. Inter-RAT, INVRAT and RNP rationales are highlighted in \hlpink{\textbf{pink}}, \hlgreen1{\textbf{green}} and \hlyellow1{\textbf{yellow}}. }
%   Examples of selective rationalizations on the charge prediction. Annotated rationales are \underline{underlined}. The rationales selected by Inter-RAT and INVRAT are highlighted in \hlpink{\textbf{pink}} and \hlgreen1{\textbf{green}} colors, respectively. The two models both predict the charge as \textit{Manslaughter} correctly, but Inter-RAT extracts more plausible rationales. More examples can be found in Appendix \hyperref[intro]{A.3}.}
%  \label{case_study}
%  \vspace{-0.4cm}
% \end{figure}
\section{Introduction}
% The remarkable success of deep neural networks has prompted the interest in applications of legal intelligence \cite{luo2017learning,zhong2018topjudge,2018interpretable,zhong2020iteratively}.
% Among them, Criminal Court View Generation has attracted increasing attention as a fundamental task of legal intelligence.
% As shown in Figure \ref{caseegg}(a), given the fact description, the goal of criminal court view generation is to yield a coherent text (i.e., court view) which is considered as a summary of the case fact and an explanation for verdicts (e.g., charges) and sentencing.
% Generating court views automatically can help lighten the workload of legal professionals (e.g. judges) and provide legal support to laymen. 

\begin{figure*}[htp]
  \centering 
  \setlength{\abovecaptionskip}{0.2cm}
  \setlength{\belowcaptionskip}{0cm}
  \includegraphics[width = 14.8cm]{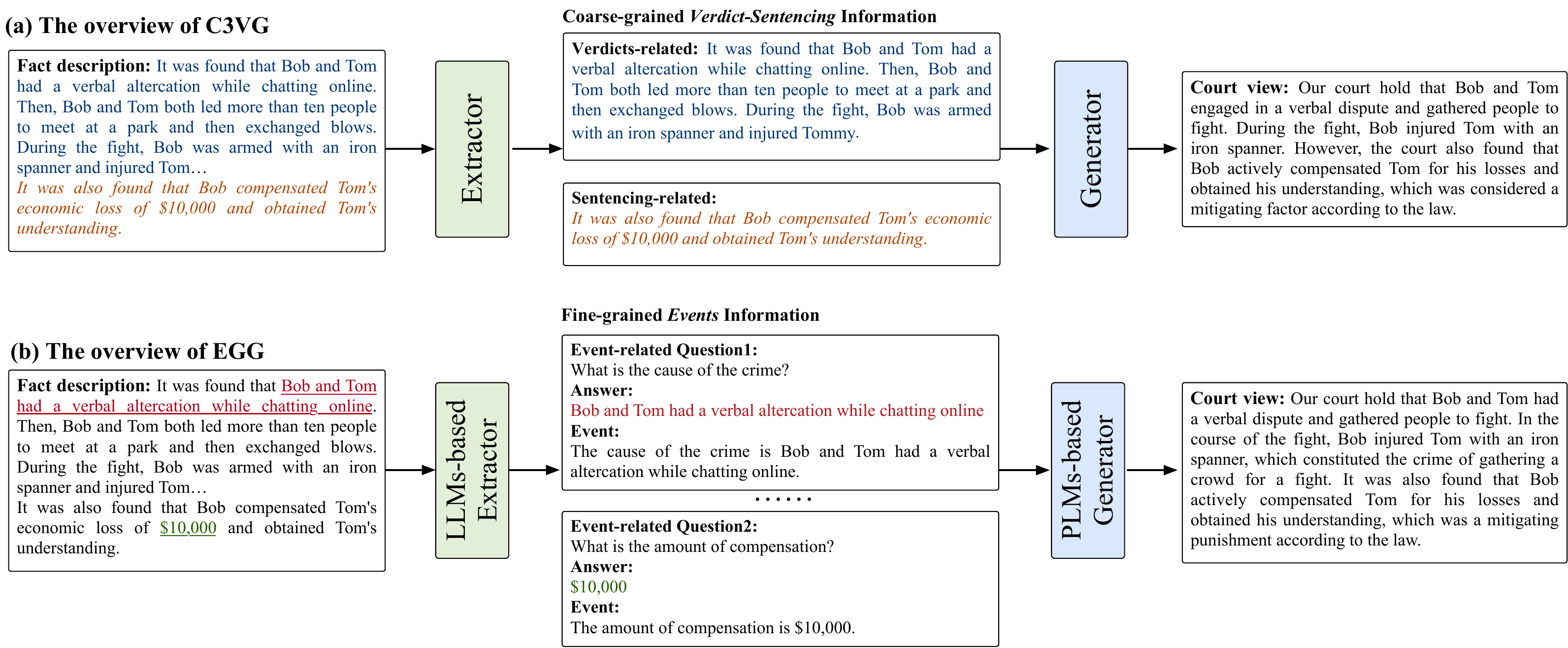}
 \caption{
  Schematic of C3VG and our method EGG presented in this paper.
 }
 \setlength{\belowcaptionskip}{0cm}
 \label{caseegg}
 \vspace{-0.4cm}
\end{figure*}

The remarkable success of deep neural networks has stimulated the exploration of legal intelligence applications  \cite{luo2017learning,zhong2018topjudge,zhong2020iteratively,wu2022towards,zhang2023contrastive,liu2023ml}. Among these applications, Criminal Court View Generation \cite{ye2018interpretable,yue2021circumstances} has garnered increasing attention as a foundational facet of legal intelligence. As depicted in Figure \ref{caseegg}(a), the objective of criminal court view generation is to produce a coherent text, referred to as a court view, which serves as a concise representation of the case facts and offers an explanation for the rendered verdicts, such as charges and sentencing. The automated generation of court views has the potential to alleviate the workload of legal professionals while providing legal assistance to laymen \cite{yue2021circumstances,wu2022towards}.

% 当前的方法分为两类，语言模型和domain-specific models。具体来说， 此外，法庭观点生成任务作为一个文本生成任务，我们可以微调大语言模型，如（Baichuan7b）来进行court view的生成。但是，从实验结果上看（表2），简单的微调并不能直接获得一个比较好的效果。这也可能是因为案件描述比较复杂，需要引入更多的领域知识。为此，一个直观的做法是将domain-specific models中的PLM替换成LLM，但是domain-specific models往往涉及到多个LM的协同训练，这对于LLM来说有很大的计算负担。另一类方法，domain-specific models，就是利用了法律领域的知识与PLM（Bart）结合来进行法庭观点生成。具体来说，

% Current approaches are divided into two categories, domain-specific models and large language models (LLMs).
% Specifically, existing domain-specific models \cite{ye2018interpretable,Huang2020Legal,yue2021circumstances} commonly generate court views based on the key information (e.g., crime circumstances \footnote{xx}) in case facts based on the legal knowledge. For example, C3VG, which achieves promising results in court view generation, explicitly groups the crime circumstances in the case facts into two coarse-grained types of circumstances (i.e., verdict-related and sentencing-related circumstances). Then, it employs the pre-trained language models (PLMs) to yield the court view based on these two type information.
% However, the components of the case facts are sufficiently complex. As shown in Figure~\ref{caseegg}(b), case facts are composed of a variety of criminal elements (aka, events), which are the underlined tokens in the fact description.
% Therefore, such a coarse-grained domain-specific method which divides the fact into two parts is unsuitable.
The existing approaches in the field can be categorized into two groups: domain-specific models \cite{ye2018interpretable,Huang2020Legal} and large language models (LLMs) \cite{touvron2023llama,ChatGPT}. Several domain-specific models \cite{yue2021circumstances,wu2022towards} commonly generate court views by leveraging key information (e.g., crime circumstances \cite{yue2021circumstances,yue2021neurjudge,yue2024circumstance}) extracted from the case facts using legal knowledge. For instance, C3VG \cite{yue2021circumstances}, a court view generation model that has demonstrated promising results, explicitly categorizes crime circumstances in the case facts into two broad types: verdict-related circumstances and sentencing-related ones. Subsequently, it employs pre-trained language models (PLMs) (e.g., BART\footnote{\url{https://github.com/yuelinan/C3VG/tree/main/bart_based_c3vg}} \cite{2020bart}) to generate court views based on these two types of information. Nevertheless, the components comprising the case facts are highly intricate. As illustrated in Figure \ref{caseegg}(b), case facts encompass various criminal elements\footnote{\url{https://en.wikipedia.org/wiki/Element_(criminal_law)}} (i.e., legal events), represented by the underlined tokens in the fact description. Consequently, the adoption of a coarse-grained domain-specific approach that partitions the facts into two segments proves to be inadequate.

Furthermore, considering that court view generation is essentially a text generation task, it is plausible to fine-tune LLMs \cite{yue2023fedjudge} (e.g., Baichuan-7B \cite{baichuan}) for court view generation. However, as evident from the experimental findings presented in Table \ref{table1}, simple fine-tuning of LLMs does not yield satisfactory results. This could be attributed to the intricacy of the fact descriptions, necessitating the incorporation of additional legal knowledge. In this regard, a straightforward approach is to substitute PLMs with LLMs in domain-specific models. Nonetheless, domain-specific models often involve the collaborative training of multiple PLMs, which poses a significant computational burden on LLMs.

To this  end, in this paper, we aim to develop a method which incorporates fined-grained event information into the court view generation by leveraging the collaboration between LLMs and PLMs in domain-specific models.
The overview of our proposed method is present in~Figure~\ref{caseegg}(b) and is two-fold: (1) extracting the fine-grained event of the
case fact and (2) generating court views based on the identified events.

However, it is a non-trivial problem.
Although available legal event extraction datasets contain substantial annotated data \cite{fengetal2022legal,yaoetal2022leven}, they primarily focus on annotating  which information belongs to events in each legal document within specific case types. This approach not only necessitates extensive professional effort but also requires re-annotation of vast amounts of legal documents when encountering new case types, thereby serving as a major bottleneck for practical applications of legal event extraction.
Therefore, it is crucial to devise a strategy that can extract events with minimal human annotation and demonstrate good generalization capabilities across different case types.
% Although existing legal event extraction datasets have massive annotated data \cite{fengetal2022legal,yaoetal2022leven}, they mainly focus on labeling each legal document in several case types. This approach not only requires extensive professional labor, but also needs to re-label massive legal documents when facing a new case type, which is the main bottleneck for applications of legal event extraction in the actual reality.
% Therefore, it is significant to design a strategy that can extract events without much human annotation and generalize well to other case types.
% Besides, since the events selected by the extractive model inevitably have omissions, the generation based solely on the extracted events may result in the loss of key information.
% Therefore,  one remaining problem is how to combine the case facts with the event information to generate more plausible and faithful court views.
%为了解决上述问题，并结合语言模型的优势，我们

To tackle the challenge mentioned above, we propose an \underline{E}vent \underline{G}rounded \underline{G}eneration (EGG) method for criminal court view generation with Cooperative (Large) Language Models following an \textit{extract}-\textit{generate} framework:

$\bullet$ \textbf{\textit{{In the extraction phase}}}, we design a \textit{LLMs}-based \textit{event extractor}. 
Specifically, we first fine-tune LLMs with the publicly available legal QA dataset CJRC \cite{duan2019cjrc} (an extractive QA dataset like SQuAD \cite{rajpurkaretal2018know}).
This fine-tuning process enables the LLMs to extract pertinent answers from the original text based on a given legal question (i.e., the prompt).
After the extractor is trained, we label each case type with several event-related questions.
For example, as shown in Figure \ref{caseegg}(b), for a case type of \textit{Mobbing}, we label the event-related questions (e.g. ``\textit{What is the cause of the crime?}'' and ``\textit{What is the amount of compensation?}''.)
Importantly,  we label these questions only for the case type itself and not for individual case facts.
When dealing with a specific case fact related to the crime of \textit{Mobbing}, we utilize the pre-defined event-related questions for the \textit{Mobbing} case type.
By prompting the trained LLMs-based event extractor with the labeled questions, we extract events for each question based on the given case fact. It is worth noting that our labeled event-related questions are not present in the CJRC dataset, thus making our extraction method a zero-shot event extraction approach.
% Notably, we label only the case type, not each case fact. For a single case fact involving the crime of \textit{Mobbing}, we use the questions labeled above for  \textit{Mobbing}.
%如图1b所示，对于一个聚众斗殴罪的案件类型，我们标注了“what is the Cause   of   the   crime”,and what is the tools of the crime.等事件相关的问题。值得注意的是，我们仅针对案件类型进行标注，而不是针对每个案件事实。对一个涉及聚众斗殴罪的案件事实，我们都使用上述为聚众斗殴罪的案件类型标注的问题。
% For instance, for a case fact associated with a particular case type (e.g., the crime of gathering a crowd for a fight), 
% Then, given a case  fact, we prompt the trained LLMs-based \textit{event extractor} to answer each labeled event question. 
% Since our labeled event -related questions do not appear in the CJRC dataset, our extraction method can be seen as a zero-shot event extraction method.
% 由于我们标注的事件问题没有出现在CJRC中，因此我们的抽取方法也是一种 zero-shot event extraction method。
% This facilitates the attainment of corresponding answers, thereby achieving 
Next, we combine the question and answer to get the events (e.g., the fine-grained events information in Figure~\ref{caseegg}(b)).
In summary, this approach only necessitates the annotation of relevant questions for each case type, with an average of 9 questions per case type. In comparison to previous methods, our proposed event extraction approach significantly reduces the annotation time required.

\begin{figure}[htp]
  \centering 
  \setlength{\abovecaptionskip}{0.3cm}
  \setlength{\belowcaptionskip}{0cm}
  \includegraphics[width = 8.cm]{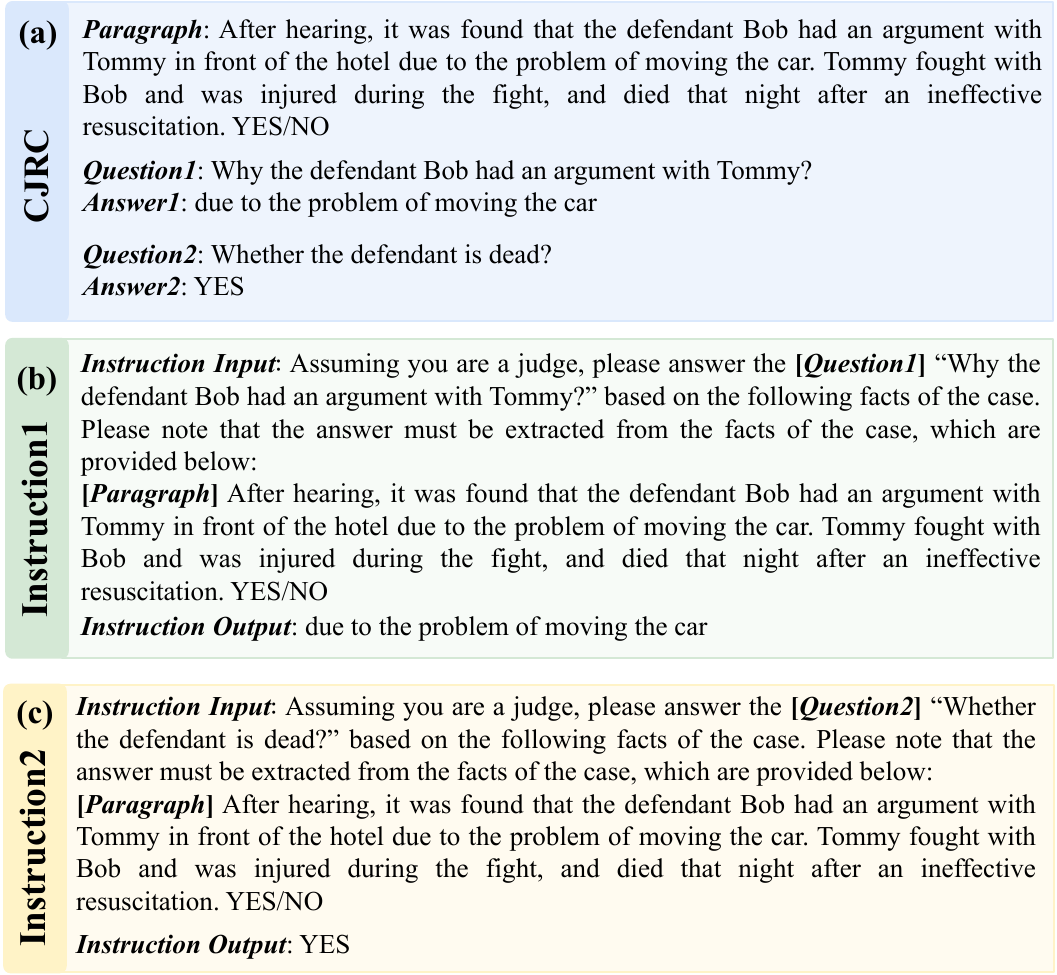}
 \caption{
  (a). An example from the CJRC dataset, including two questions. 
  (b). An example from the processed instruction dataset 
based on the \textit{{Question1}} in (a).
(c). An example from the processed instruction dataset 
based on the \textit{{Question2}} in (a).
 }
 \label{crimeexample}
 \vspace{-0.5cm}
\end{figure}
$\bullet$ \textbf{\textit{{In the generation phase}}}, we splice the facts and events together to form a new text input, which is then fed into the \textit{PLMs}-based generator to yield the court views.

Additionally, taking into consideration the computational burden posed by the use of LLMs in the extraction phase of EGG, we recognize the need to enhance its practical applicability for both laymen and professionals. To address this, we propose an LLMs-free EGG method, referred to as $\textmd{EGG}_{free}$, which eliminates the requirement of events during the inference phase. 
Specifically, in the training process, we still employ LLMs to extract events in the extraction phase. However, in the generation phase, instead of merging the event and fact as input to the generator, we leverage the event as auxiliary information to assist the model in generating the court view based solely on the fact. To achieve this, we encode the fact and event separately using the fact and event encoders. Subsequently, we design a contrastive learning module to facilitate the fact encoder in capturing co-occurrence signals with the event through contrastive constraints. Finally, we generate the court view based on the fact. 
Importantly, during the inference phase, $\textmd{EGG}_{free}$ no longer requires any event information. It solely relies on the case fact to generate the court view without the need for event extraction. This modification aims to improve the practicality and usability of the EGG method for legal professionals and individuals without legal expertise.
%此外，考虑到我们的模型在抽取阶段使用到了LLM，这增加了模型推理时的计算负担，使得模型在实际应用中受限。为了使我们的模型可以更好的辅助layman和专业人士，我们提出了一个在推理阶段llm-free的方法，命名为eggfree。具体来说，在训练过程中，在抽取阶段，我们依然使用llm抽取出事件。在生成阶段，不用于之前的egg将event和fact进行合并作为生成器的输入，我们将event作为一种辅助信息，辅助模型根据fact来生成法庭观点。具体来说，我们分别将fact和事件通过事实和事件编码器进行编码，之后我们设计来一个对比学习模块来teach the fact encoder to capture co-occurrence signals with event through contrastive constraints. It is important to note that during the inference phase, eggfree does not require any event. It solely relies on the case fact to generate  the court view  without the event extraction. 

In summary, the major contributions of this paper are:
\begin{itemize}[leftmargin=*]
  \item We propose an Event Grounded Generation (EGG) method for criminal court view generation with cooperative (large) language models, which \textit{first} introduces the fine-grained event information into the court view generation.
  \item We propose a low data resource approach to achieve a zero-shot legal event extraction with LLMs.
  \item To alleviate the computational burden in EGG during inference that employs LLMs, we propose a LLMs-free EGG method based on the contrastive constraint.
  \item Extensive experiments on a real-world dataset validate the effectiveness of our method by comparing it with several competitive~methods.
\end{itemize}
%1.提出了融合法律事件的法庭观点方法。with协同的语言模型，首次在法庭观点生成中引入事件信息。2.我们提出了一个低数据资源的法律事件抽取方法，zero-shot QA for event extraction。3.为了缓解使用LLM的计算负担in egg during 推理，我们提出了一个基于对比学习的free方法。4.实验结果证明了我们提出方法的有效性

% Besides, ROUGE \cite{lin2004rouge} and BLEU \cite{papineni2002bleu} metrics have been employed to evaluate court views in many researches \cite{ye2018interpretable,yue2021circumstances}. Since such string matching metrics fail to capture the semantics of the generated court views sufficiently, we propose a new metric PRISON SCORE to evaluate the generated court views.
% Specifically, we first adopt the generated court view and the gold one to predict the term of penalty (a downstream task of the court view generation), respectively. Then, we calculate the difference between their scores, and the smaller the difference, the closer we consider the generated court view to the true one.

\section{Related Work}
\textbf{Court View Generation.}
The remarkable success in neural networks provokes the legal intelligence \cite{zhong2018topjudge,zhong2020iteratively,shao2021investigating,ma2021lecard,13591651,103591631}. Among them, court view generation has achieved increasing attention \cite{ye2018interpretable,yue2021circumstances}.
Specifically, 
\cite{ye2018interpretable} were the first to formulate the task of court view generation and explored the use of charges to enhance the generation process, allowing the model to focus on verdict-related information within the case facts.
\cite{Huang2020Legal} proposed a court view generation approach that involved masking key tokens in a template and subsequently employing a question-answering (QA) method to fill in these masked tokens.
\cite{wu2022towards} integrated legal judgment prediction with court view generation, enabling the simultaneous generation of judgment results and court views.
\cite{yue2021circumstances} designed an extract-generate framework that categorized case facts into two types, namely verdict-related and sentencing-related information, using an extractor. The generated court views were then based on the extracted information.
Despite the promising results achieved by these existing methods, they have overlooked the incorporation of fine-grained event information present in case facts. This limitation highlights the need to consider and leverage event information for more comprehensive and accurate court view generation.

\textbf{Large Language Model in Legal AI.}
Large Language Models (LLMs) such as ChatGPT \cite{ChatGPT} and LLaMA \cite{touvron2023llama} have exhibited impressive performance across various complex tasks and have made a significant impact on society. In the realm of legal AI, researchers have been combining LLMs with legal tasks \cite{wang2023pangu,yue2023disc,HanFei}.
One notable example is Lawyer LLaMA \cite{lawyerllamareport}, which underwent continual pretraining on an extensive legal corpus to systematically acquire legal knowledge. The model was then fine-tuned using legal instruction data, enabling it to apply its legal knowledge to specific scenarios. This approach leverages the power of LLMs to enhance the effectiveness of legal AI tasks.
Another approach, ChatLaw \cite{cui2023chatlaw}, explored the use of larger base models to improve the logical reasoning capabilities of legal models. By leveraging the increased capacity and capabilities of larger models, ChatLaw aimed to enhance the model's ability to perform complex legal reasoning tasks.
Privacy concerns in the legal domain are addressed by FedJudge \cite{yue2023fedjudge}, which adopts Federated Learning during the instruction tuning process. This approach ensures the privacy of legal data by training the model on local devices and only sharing aggregated updates, rather than sharing raw data.
In this paper, the focus is specifically on utilizing LLMs to achieve legal event extraction. 
\begin{figure*}[htp]
  \centering 
  \setlength{\abovecaptionskip}{0.2cm}
  \setlength{\belowcaptionskip}{0cm}
  \includegraphics[width = 13.7cm]{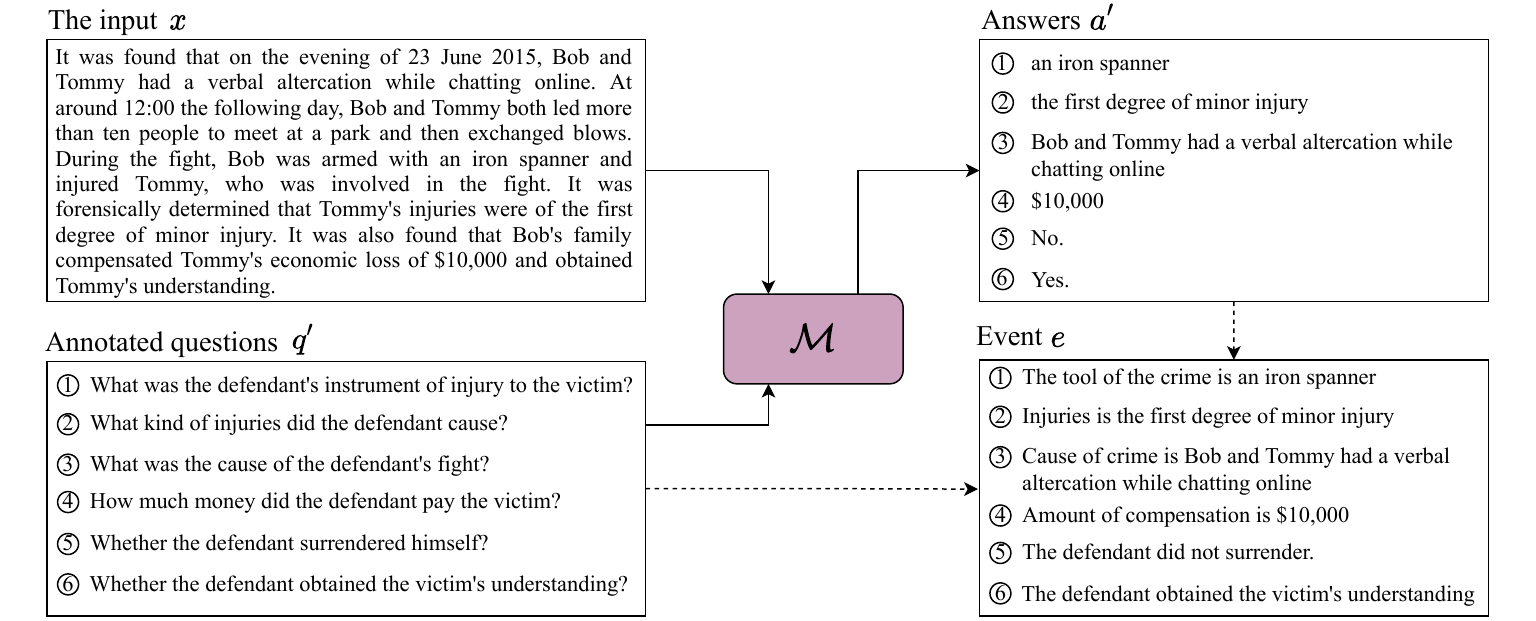}
 \caption{
  The process of event extraction. Among them, the input $x$ is a legal document in the ``crime of \textit{Affray}'' case type, and the questions $q^{\prime}$ is annotated for 
  the ``crime of \textit{Affray}'' case type (not only for the input $x$).
  % ``\includegraphics[width=20pt]{figure/real.png}'' is the process of model inference and 
  % ``\includegraphics[width=20pt]{figure/notreal.png}'' represents the post-processing process. 
 }
 %\vspace{-0.3cm}
 \label{model}
 
\end{figure*}

\textbf{Legal Event Extraction.}
In the field of legal event extraction, numerous studies \cite{3345608,shenetal2020hierarchical,9194466,yaoetal2022leven,fengetal2022legal} have involved annotating legal event types for each legal document. Notably, \cite{yaoetal2022leven} have annotated over 8,000 legal documents with 108 event types. However, this manual annotation process is labor-intensive and time-consuming. Furthermore, when encountering new legal event types, it becomes necessary to label additional data, making existing datasets less reliable.
It is crucial to develop an event extraction method that minimizes the reliance on extensive manual annotation.

\section{Event Grounded Generation for Criminal Court View}
% In this section, we first formulate the criminal court view generation task. Then, we present the details of our EGG, consisting of an \textit{event extractor} and a \textit{court view generator}.

\subsection{Problem Definition}
Here, we explore the problem of criminal court view generation. We first clarify the definitions of the terms as follows:

\textbf{Fact description} $x=\left\{x_{1}, x_{2}, \ldots, x_{n}\right\}$ is the identified facts in a case including several events, where $x_{i}$ denotes the \textit{i}-th token.

\textbf{Event set} $e=\left\{e_{1}, e_{2}, \ldots, e_{k}\right\}$ consists of $k$ events of the fact, where $e_{i} = \left\{e_{i}^{1}, e_{i}^{2}, \ldots, e_{i}^{m}\right\}$ contains $m$ tokens and each event is a subsequence of the fact.

\textbf{Court view} is the summary of the fact which consists of the charge $c$ and rationales $r=\left\{r_{1}, r_{2}, \ldots, r_{t}\right\}$. Among them, the rationale is concluded from the fact in order to determine and support the judgment results, such as sentencing.
In this work, we assume the charge is available, and we only focus on generating rationales in court views, where the charge can be easily obtained by the judge or the charge prediction systems~\cite{zhong2018topjudge,yue2021neurjudge,zhang2023contrastive}.

Then, based on the above definitions, our problem is defined as:

\textbf{Problem 1} (Court View Generation).
\textit{Given the case fact $x$, our goal is first to extract several events~$e$ from the case fact, and then generate the rationales $r$ in court views, where the gold events are unavailable.}

\subsection{Architecture of EGG}
Our proposed Event Grounded Generation (EGG) for criminal court view method consists of two phases, cascading the \textit{event extractor} and the \textit{court view generator}.
Specifically, in the extraction phase, we first train a LLMs-based QA model which can extract a subsequence of the text input as the answer to the prompts (or questions).
After the model is trained, we consider this model as the \textit{event extractor} to select several events from the case fact by introducing annotated legal  event-related questions.
Finally, we employ a PLMs-based \textit{court view generator} to generate court views by merging the fact and event as the new text input.

\subsubsection{\textbf{Event extractor}}
Existing legal event extraction datasets \cite{yaoetal2022leven,fengetal2022legal} mainly focus on annotating each case under different case types (i.e., different charges). However, this annotation requires significant and expensive professional labor. Meanwhile, when facing a new case type, it  commonly needs to be re-labeled. To this end, we develop a zero-shot LLMs-based legal event extractor.

Specifically, we implement the \textit{extractor} with a publicly available legal QA dataset CJRC \cite{duan2019cjrc}, which consists of the paragraph $p$, question $q$ and answer $a$ as shown in Figure \ref{crimeexample}(a). Among them, the answer $a$ is a part of paragraph~$p$. 
Besides, to answer the question about YES or NO, CJRC adds ``YES/NO'' at the end of the paragraph.
Based on CJRC, we train a legal LLMs-based model $\mathcal{M}$ to extract answers from the paragraph according to the questions. 

In detail, we begin by transforming the original CJRC dataset into an instruction dataset \cite{alpaca}, denoted as $\mathcal{D}$, where each instruction data has the form of $\{Instruction Input: Instruction Output\}$. Figure \ref{crimeexample}(b) and Figure \ref{crimeexample}(c) illustrate the specific format of the prompt, task-specific instruction, and ground truth in our instruction dataset.
Next, we utilize the instruction tuning method to fine-tune the base generative LLMs to extract answers  from paragraphs. To address the computational and time constraints associated with directly fine-tuning the entire LLM, we employ the parameter-efficient fine-tuning technique for training the extractor.
Specifically, we employ the LoRA \cite{hu2022lora} method which involves freezing the pre-trained model parameters and introducing trainable rank decomposition matrices into each layer of the Transformer architecture \cite{vaswani2017attention}.
Finally, the learning objective can be computed as:
\begin{equation}
  \mathcal{L}_{e}=- \sum_{t=1}^{|y|} \log \left(P_{\Theta+\Theta_{L}}\left(y_{t} \mid m, y_{<t}\right)\right),
  \label{eq1}
\end{equation}
% \begin{equation}
%   \mathcal{L}_{e}=-\sum_{(m, y) \in \mathcal{D}} \sum_{t=1}^{|y|} \log \left(P_{\Theta+\Theta_{L}}\left(y_{t} \mid m, y_{<t}\right)\right),
%   \label{eq1}
% \end{equation}
where $m$ and $y$ represent the \textit{Instruction Input} and \textit{Instruction Output}, $y_{t}$ denotes the $t$-th token of $y$, $y_{<t}$ is the tokens before $y_{t}$, and $\Theta$ represents the frozen LLMs parameters and $\Theta_{L}$ is the trainable LoRA parameters ($\Theta_{L} \ll \Theta$).

After the LLMs-based extractor is trained, we annotate several questions for each case type, where each question $q^{\prime}$ is related to the event in case facts $x$.
It is important to note that we label the questions only for the case type and not for each individual case fact. For instance, if we have a case fact related to the crime of \textit{Affray}, we utilize the previously labeled questions for \textit{Affray}, such as the cause of the crime, tools of the crime, and whether to surrender.
Then, as shown in Figure~\ref{model},
we promote the trained LLM $\mathcal{M}$ to answer these event-related questions and obtain the corresponding answers $a^{\prime}$ (i.e.,  $a^{\prime} = \mathcal{M}(x,q^{\prime})$). Finally, we combine the obtained answers $a^{\prime}$ and the corresponding questions $q^{\prime}$ to obtain the event $e$.
\begin{figure}[htp]
  \centering 
  \setlength{\abovecaptionskip}{0.3cm}
  \setlength{\belowcaptionskip}{0cm}
  \includegraphics[width = 8.cm]{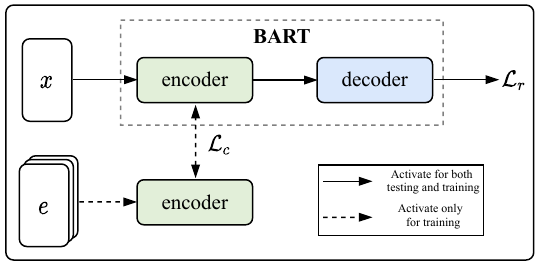}
 \caption{
  Architecture of $\textmd{\textbf{EGG}}_{\textbf{free}}$.
 }
 \label{eggfree}
 \vspace{-0.6cm}
\end{figure}
It is worth emphasizing that the labeled event-related questions used in this process are not present in the CJRC dataset. This characteristic distinguishes our extraction method as a zero-shot event extraction approach, as it successfully extracts event information from case facts using questions that were not available in the training data. 
This zero-shot capability increases the versatility and adaptability of our approach to handle new or unseen case types.

\subsubsection{\textbf{Court view generator}}
Previous models generate court views based solely on case fact. In this section, our \textit{court view generator} designs a strategy to incorporate extracted event information into the fact to yield more plausible court views, where we adopt the BART \cite{2020bart} as our backbone by considering the advantages of the current PLMs. Specifically, we  merge the event and fact descriptions to form new input $x^{\prime}$ of the \textit{court view generator}.
In practice, limited by the maximum length of the PLMs, we enforce the events to be placed before facts (i.e., $x^{\prime} =  e ||| x $), where ``$|||$'' represents the process of mergers.
% First, from the model architecture side, we introduce an adaptive incorporation method.
% Specifically, we pass the fact $x$ and event $e$ into the BART to obtain their corresponding representations $h^{x}_{t} \in \mathbb{R}^{d}$ and $h^{e}_{t} \in \mathbb{R}^{d}$ at each step $t$ of rationales $r$ in the court views.
% Inspired by \cite{cao2018faithful}, different from directly exploiting these representations to yield the probability of rationale tokens~$r_{t}$, we employ an adaptive method with a gate network:
% \begin{equation}
%   \begin{aligned}
%     g_{t} &= \operatorname{sigmoid}\left( {W}_{h}[h^{x}_{t};h^{e}_{t}]\right), \\
%     h_{t} &= g_{t} \odot h^{x}_{t} + (1-g_{t})\odot h^{e}_{t}, \\
%     P\left(r_{t} | r_{<t}\right)&=\operatorname{softmax}\left({W}_{r} {h}_{t}\right),
%   \end{aligned}
%   \label{gate}
%   \end{equation}
% where the gate value $g_{t} \in [0,1]$,  ${W}_{h} \in \mathbb{R}^{1 \times 2d}$, and ${W}_{r} \in \mathbb{R}^{|V| \times d}$. Among them, $|V|$ is the length of vocabulary.
% Then, we denote this method as $\textmd{EGG}_{adaptive}$.
% and at each step t, the probability to v predict y t a is computed as follows
% Different introducing the event information into the model architecture, we merge the event and fact descriptions to form new input $x^{\prime}$ of \textit{court view generator}.
% In practice, limited by the maximum length of the pre-trained model, we enforce the events to be placed before facts (i.e., $x^{\prime} =  e ||| x $), where ``$|||$'' represents the process of mergers. We name this method as EGG.

\subsection{Training and Inference}
In this section, we describe the training loss in our proposed method EGG. Specifically,
in the extraction phase, we employ Eq(\ref{eq1}) to train our LLMs-based event extractor.
In the generation phase, we adopt the negative log-likelihood loss to optimize the generator:  
% \begin{equation}
%   \mathcal{L}_{r}=\frac{1}{T} \sum_{t=1}^{T} -\log P\left(r_{t}\right),
%   \label{loss_short}
% \end{equation}
\begin{equation}
  \mathcal{L}_{r}=- \sum_{t=1}^{|r|} \log \left(P_{\Theta_{B}}\left(r_{t} \mid x^{\prime}, r_{<t}\right)\right),
  \label{loss_short}
\end{equation}
where $\Theta_{B}$ is the trainable BART parameters, $r_{t}$ denotes the $t$-th token of $r$ and $r_{<t}$ is the tokens before $r_{t}$.

During the inference phase, given a description of case fact, we first use the LLMs-based \textit{event extractor} to extract the events from the case fact. Then, we generate the court view based on both the facts and events.

\section{$\textmd{\textbf{EGG}}_{\textbf{free}}$: LLMs-free EGG with contrastive constraints}
Indeed, the use of LLMs for event extraction in the extraction phase of EGG can lead to increased computational burden during the inference phase. This limitation hampers the practical application of the model in real-world scenarios. To overcome this challenge, we propose an LLMs-free EGG method that employs contrastive constraints, enabling court view generation without the need for event information during the inference phase.

\subsection{Architecture of $\textmd{\textbf{EGG}}_{\textbf{free}}$}
During training, $\textmd{EGG}_{free}$ follows the \textit{extractor}-\textit{generator} framework.
Specifically, in the extraction phase, similar to EGG, we still employ LLM to extract events. 
In the generation phase, unlike the previous EGG of combining event and fact as inputs to the generator, we use event as a kind of auxiliary information to assist the model in generating court views based on fact. In particular, as shown in Figure \ref{eggfree}, given the fact and event, we first employ the fact encoder and event encoder to encode both fact $x$ and event $e$ as the corresponding representations $h_{x} \in \mathbb{R}^{d}$ and $h_{e} \in \mathbb{R}^{d}$, where $d$ is the dimensional size. 
Then, we feed the fact representation into the decoder for court view generation. In practice, we use the encoder and decoder of  BART to achieve the above implementation. 

Subsequently, to enable the fusion of event information into $\textmd{EGG}_{free}$, we employ a novel contrastive learning strategy during the training phase. This strategy aims to teach the fact encoder to memorize the co-occurrence event signals within its parameters, allowing the fact encoder to inject event clues into fact representations during the inference phase.

In particular, during the training phase, as shown in Figure \ref{eggfree}, we adjust the parameters of the fact encoder based on the event encoder to maximize the mutual information between the case fact and event. To achieve this objective, for a training fact  representation $h_{x}$, we build its positive sample set using its corresponding event $h_{e}$ (referred to $h_{e}^{+}$), i.e., $\mathcal{N}^{+}=\{h_{e}^{+}\}$,
and its negative sample set $\mathcal{N}^{-}=\mathcal{N}_{batch}\char92\mathcal{N}^{+}$ where $\mathcal{N}_{batch}$ denotes each batch of event samples.
To teach the fact encoder to memorize the co-occurrence event signals, we define the contrastive loss following the concept of InfoNCE~\cite{oord2018representation}. The contrastive loss is formulated as:
\begin{equation}
    \begin{split}
        \mathcal{L}_{c}=-\mathbb{E}_{h_{e}^{+}\in\mathcal{N}^{+}}\left[
        \log\frac{\exp{(\frac{\operatorname{sim}\left(h_{x}, h_{e}^{+}\right)}{\tau})}}{\exp{(\frac{\operatorname{sim}\left(h_{x}, h_{e}^{+}\right)}{\tau})}+\sum_{h_{e}^{-}\in\mathcal{N}^{-}}\exp{(\frac{\operatorname{sim}\left(h_{x}, h_{e}^{-}\right)}{\tau})}}
        \right],
    \end{split}
    \label{eq:contrastive_loss}
\end{equation}
where $\text{sim}(h_{x}, h_{e})$ represents the similarity measure between the fact representation $h_{x}$ and the event representation $h_{e}$, and $\tau$ is a temperature parameter that controls the sharpness of the probability distribution.
Besides, we set the fact encoder and the event encoder to share parameters to save GPU memory. According to our experiments, separate encoders and shared encoders do not have a significant difference in the generation performance.

%我们依然使用llm抽取出事件。在生成阶段，不用于之前的egg将event和fact进行合并作为生成器的输入，我们将event作为一种辅助信息，辅助模型根据fact来生成法庭观点。具体来说，我们分别将fact和事件通过事实和事件编码器进行编码，之后我们设计来一个对比学习模块来teach the fact encoder to capture co-occurrence signals with event through contrastive constraints. It is important to note that during the inference phase, eggfree does not require any event. It solely relies on the case fact to generate  the court view  without the event extraction. 

\subsection{Training and Inference}
In the training process, since $\textmd{EGG}_{free}$ uses the same extractor as EGG, we use Eq(\ref{eq1}) to train the extractor. Besides, the final objective of the generator in $\textmd{EGG}_{free}$ is defined as:
\begin{equation}
  \mathcal{L}_{\textmd{EGG}_{free}} =  \mathcal{L}_{r} + \beta \mathcal{L}_{c},
\end{equation}
where $\beta$ is the adjusted hyperparameter.

During the inference phase, since the fact encoder learns to capture co-occurrence signals with the event through contrastive constraints, $\textmd{EGG}_{free}$ can ignore the event as the input, enabling the generation of contextually relevant court views based solely on the case fact. 

By leveraging contrastive constraints, our proposed method eliminates the reliance on LLMs for event extraction in the inference phase. This approach significantly reduces the computational burden, making the model more suitable for real-world applications. 
\begin{table}[htp]

  \centering
  \setlength{\abovecaptionskip}{0.1cm}
  \setlength{\belowcaptionskip}{0pt}
  \caption{The statistics of datasets.}
  \setlength{\abovecaptionskip}{0.1cm}
\setlength{\belowcaptionskip}{0pt}
  \renewcommand\arraystretch{1}
  \setlength{\tabcolsep}{2.mm}{
    \scalebox{.82}{
  \begin{tabular}{ll}
    
    \toprule
      \text {\bfseries CJO } & \text {\bfseries Results } \\
      \midrule
      \text{\# Sample} & 62,939\\
      \text{\# Types of cases} & 62\\
      \text{\# Avg. Length of fact description} & 458.1\\
      \text{\# Avg. Length of court view description} & 130.9\\
      \text{\# Avg. Annotated questions of event} & 9.0\\
      \text{\# Avg. Length of annotated questions of event} & 10.8\\
      \text{\# Avg. Length of all events in a case } & 83.4\\
      \bottomrule
      \toprule
      \text {\bfseries CJRC } & \text {\bfseries Results } \\
      \midrule
      \text{\# Sample} & 20,000\\
      \text{\# Avg. Length of paragraph} & 501.8\\
      \text{\# Avg. Length of question} & 16.5\\
      \bottomrule
  \end{tabular}}}
  \label{table2}
\end{table} 

\section{Experiments}

To evaluate the effectiveness of EGG, we conduct experiments to answer the following research questions:
% 主实验，刑期预测的下游任务，长度实验，效率比较，对比学习的tsne，可视化实验
\begin{itemize}[leftmargin=*]
  \item \textbf{RQ1:} How effective are EGG and $\textmd{EGG}_{free}$ in improving the performance of event extraction and court view  generation?
  \item \textbf{RQ2:} How efficient is $\textmd{EGG}_{free}$ during the inference phase?
  \item \textbf{RQ3:} What are the performances of EGG by the length of court views?
\item \textbf{RQ4:} How do EGG and $\textmd{EGG}_{free}$ perform in human evaluation?
% \item \textbf{RQ5:} Can the contrastive constraint in $\textmd{EGG}_{free}$ teach the fact encoder to memorize the co-occurrence event signals?
\item \textbf{RQ5:} What is the court view generated by EGG  to a specific case~fact?

\end{itemize}

\subsection{Datasets}
In the extraction phase, we adopt the criminal cases in CJRC\footnote{\url{https://github.com/china-ai-law-challenge/CAIL2019}} \cite{duan2019cjrc} as the training data, where we process CJRC into the format of an instruction dataset.
Figure \ref{crimeexample} is an example from the CJRC and the instruction dataset.
In the generation phase, following \cite{yue2021circumstances}, we conduct experiments on CJO\footnote{\url{https://github.com/bigdata-ustc/C3VG}}, where CJO is collected from the published legal documents in China Judgments Online\footnote{\url{https://wenshu.court.gov.cn}}.
% Besides, we also design a new metric to evaluate the generated court views.
Detailed dataset statistics are shown in Table \ref{table2}.
Among them, since there exist 62 types of cases, we ask three law expects to annotate questions for each case type, for a total of 558 questions.

\subsection{Experimental Setup}
In this section, we present the detailed experimental setup of our proposed EGG. First, in the extraction phase, we adopt Baichuan-7B \cite{baichuan} as the backbone of the LLMs-based event extractor $\mathcal{M}$. Then, we employ the LoRA to parameter-efficient fine-tune it on the instruction dataset. 
For training, we adopt an AdamW optimizer \cite{LoshchilovH19} with an initial learning rate of 1e-5, then we set the maximum sequence length as 512 and the batch size as 4. Besides, the rank of LoRA is set to 4.
In the generation phase, we employ BART \cite{2020bart} to generate the court views. We set the learning rate to 1e-4 and the batch size to 8, and $\beta$ in  $\textmd{EGG}_{free}$  to 1.
For evaluation, we adopt macro-average F1 as our metric to evaluate the performance of the LLMs-based event extractor $\mathcal{M}$ in the test set of CJRC.
Besides, since there exist no gold events in CJO, we assume that the better the generated court views perform, the more effective events are extracted.
To this end, to evaluate the performance of the generation, we adopt ROUGE \cite{lin2004rouge} and BLEU \cite{papineni2002bleu}  as the metrics. Among them, we report F1 scores of ROUGE-1, ROUGE-2, and ROUGE-L, and we keep the result of BLEU-1, BLEU-2 and BLEU-N (i.e., an average score of BLEU-1, BLEU2, BLEU-3, and~BLEU-4).
\begin{table*}[htbp]
  \setlength{\abovecaptionskip}{0.1cm}
  \renewcommand\arraystretch{0.8}
  \centering
  \caption{Results of court view generation. ``*'': results obtained from C3VG \cite{yue2021circumstances}.}
  \label{results}
  \setlength{\tabcolsep}{5.2mm}{
      \scalebox{0.8}{
        \begin{tabular}{c|ccc|ccc|c}
          \hline    
          \toprule
           \multirow{2}{*}{Models}  
           &\multicolumn{3}{c}{ROUGE ($\uparrow$) } & \multicolumn{3}{c|}{BLEU ($\uparrow$)} &  \multirow{2}{*}{Bert-S ($\uparrow$) }   \\
           \cmidrule(r){2-4} \cmidrule(r){5-7} 
           &R-1&R-2&R-L&B-1&B-2&B-N  \\  
          \midrule

         $\textmd{AttS2S}^{*}$         & 58.7 & 38.9 & 59.4 & 50.5 & 41.0 & 38.0 & --\\
              $\textmd{PGN}^{*}$                & 59.3 & 37.0 & 59.8 & 50.2 & 39.6 & 36.7 &--\\
               $\textmd{Transformer}^{*}$       & 59.9 & 39.6 & 60.9 & 50.8 & 41.3 & 38.1 &--\\
             $\textmd{Label-AttS2S }^{*}$       & 47.0 & 31.4 & 52.8 & 38.7 & 31.6 & 29.4 &--\\
                $\textmd{C3VG}^{*}$             & 60.1 & 40.5 & 62.5 & 52.1 & 43.5 & 40.6 &--\\
            \midrule
            BART(Event)     & 66.04 & 50.46 & 54.01 & 52.16 & 47.95 & 46.76 &81.93  \\
            BART(Fact)       & 74.96 & 58.34 & 62.44 &  54.94 & 52.55 & 51.46 &85.59 \\
            C3VG with BART    & 75.59  & 64.22 & 65.11 & 56.16 & 53.58 & 52.71 & 85.61\\

                \midrule
              Baichuan-7B & 57.43 & 37.76 & 38.65 & 58.96 &  53.92 & 52.01& 72.84\\
              Baichuan-7B(Fact) &  74.05 & 60.25 &  61.69& \textbf{69.48} & \textbf{66.02} &  \textbf{64.71} &82.58 \\

                \midrule

               $\textmd{EGG}_{free}$ & 75.23 & 64.41& 64.24 & 57.19& 54.34& 53.41 &85.86\\
                $\textmd{EGG}$ & \textbf{76.86} & \textbf{65.15} & \textbf{65.90} & 56.92 & 54.43 &  53.59 & \textbf{86.54}\\
          \bottomrule
          \hline

      \end{tabular}

      }
  }  
  \label{table1}
\end{table*}

\subsection{Comparison methods}
In this section, to evaluate the generated court view, we employ three type of baselines. First, we compare EGG with several traditional baselines:
\begin{itemize}[leftmargin=*]
  \item \textbf{AttS2S} \cite{bahdanau2014neural} is an attention-based sequence-to-sequence model, following an encoder-decoder framework.
  \item \textbf{PGN} \cite{SeeLM17} employs a pointer network to solve the out of vocabulary (OOV) problem in the text generation.
  \item \textbf{Transformer} \cite{vaswani2017attention} has been widely implemented to generate texts.
  \item \textbf{Label-AttS2S} \cite{ye2018interpretable} is designed to generate court views by introducing the charge semantics into AttS2S.
  \item \textbf{C3VG} \cite{yue2021circumstances} separates the case fact into two parts with an \textit{extract}-\textit{generate} framework to generate the court views.
\end{itemize}

The above baselines are implemented with GRU \cite{cho2014learning} or transformer.
For a fair comparison, the results of the above baselines are directly taken from \cite{yue2021circumstances}.

Besides, since the pre-training models have promoted the text generation in recent years, we introduce several approaches based on the pre-training models:
\begin{itemize}[leftmargin=*]
  \item \textbf{BART} \cite{2020bart} is a Transformer-based pre-training sequence-to-sequence model, which achieves promising results in text generation.
  In this paper, BART(Fact) denotes BART takes the case fact as the input. BART(Event) represents taking the extracted event as the input.
  \item \textbf{C3VG with BART} \cite{yue2021circumstances} implements C3VG with BART as the backbone.
\end{itemize} 
Finally, we also compare LLMs baselines with EGG:
\begin{itemize}[leftmargin=*]
  \item \textbf{Baichuan-7B} \cite{baichuan} is a large language model which  achieves competitive results in Chinese intelligence tasks.
  \item \textbf{Baichuan-7B(Fact)} employs LoRA to fine-tune Baichuan-7B by taking the case fact as the input  with the form of the instruction dataset. Among them, the \textit{Instruction Input} is: ``Assuming you are a judge, please  summarize the facts of the case: [the description of case facts]'', and  the \textit{Instruction Output} is the court views.
\end{itemize}

\subsection{Performance on Event Extraction and Court View Generation (RQ1)}
\subsubsection{\textbf{Results of event extraction}}
In this section, we report the macro-average F1 to evaluate the performance of the LLMs-based extraction model $\mathcal{M}$ in the test set of CJRC. After statistics, the macro-average F1 is \textbf{\underline{84.6}} which performs better than the original macro-average F1 (82.9) reported in the paper of CJRC \cite{duan2019cjrc} which employs BERT \cite{kenton2019bert} to achieve the answer extraction. This observation demonstrates the effectiveness of LLMs on extraction.
However, our goal is employing the trained $\mathcal{M}$ to predict the potential events in our court view data CJO. Therefore, the F1 scores in the test set of CJRC fail to illustrate the effectiveness of our extraction sufficiently.
To further evaluate the extracted events,
we consider that the better the generated court view performs, the more effective the extracted events will be. The corresponding results are shown in section  \ref{4.4.2}.

\begin{figure*}[htp]
  \centering 
  \setlength{\abovecaptionskip}{0.1cm}
  \setlength{\belowcaptionskip}{0cm}
  \includegraphics[width = 15.7cm]{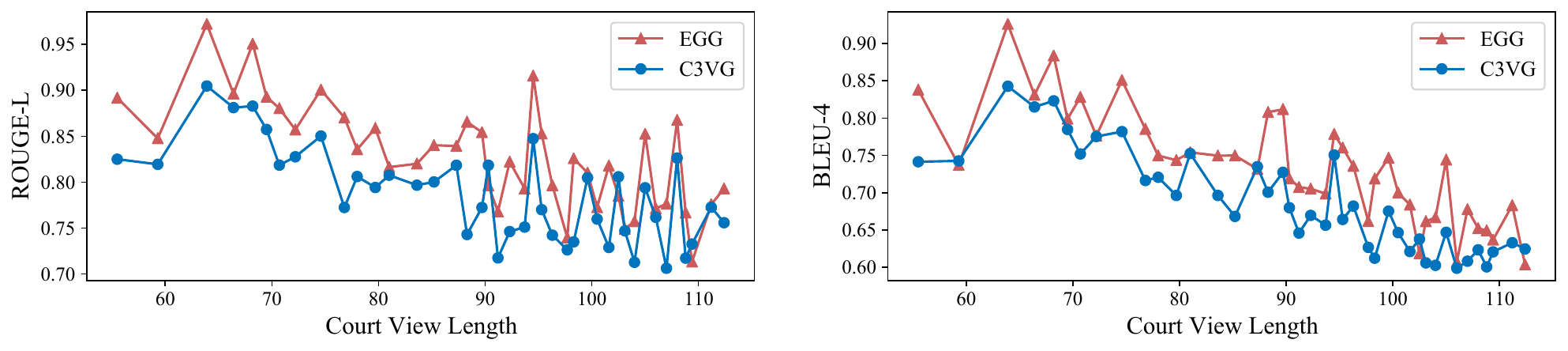}
 \caption{
  Model performance by the length of court views. Among them, C3VG is implemented with BART.
 }
 \label{rouge}
 %\vspace{-0.7cm}
\end{figure*}

\subsubsection{\textbf{Results of court view generation}}
\label{4.4.2}
To validate the effectiveness of EGG, we first compare it with several baselines. As shown in Table \ref{table1}, we find all methods that exploit PLMs outperform the traditional baselines implemented with GRU or transformer, which demonstrates the effectiveness of PLMs. 
Then, we can observe that both $\textmd{EGG}$ and $\textmd{EGG}_{free}$ perform better than other baselines in most metrics, which indicates our EGG can generate more plausible court views.
Specifically, compared with C3VG with BART, which groups the original fact into two type paragraphs to generate court views, our EGG significantly outperforms it. This observation demonstrates that incorporating fine-grained events into court view generation is more effective than employing coarse-grained paragraphs.
We also implement BART by taking fact and event as the text input, respectively.
From the results, we observe that BART(fact) surpasses  BART(Event) by a large margin, illustrating that there exist several events are not extracted. These observations prove that it is necessary to design an incorporated strategy to combine the case facts with the event information to generate court views.

% \begin{table}
%   \setlength{\belowcaptionskip}{.1cm}
%   \setlength{\abovecaptionskip}{0.1cm}
%   \caption{Inference speed and model parameters.}
%   \centering
%   % \setlength{\belowcaptionskip}{.0cm}
%   % \setlength{\abovecaptionskip}{0.cm}
%   \renewcommand\arraystretch{.9}
%   \setlength{\tabcolsep}{6mm}{
%   \scalebox{0.85}{
%         \begin{tabular}{c|cc}
%            \hline    
%            \toprule
%               Methods&Inference Speed&Parameters\\
%            \midrule
%            Baichuan-7B(Fact)       & 1.00 $\times$ &  7B \\
           
%               \midrule
%               $\textmd{EGG}_{free}$  & 0.71 $\times$ &  139M \\
%               EGG  & 8.72 $\times$ &  139M + 7B \\
%            \bottomrule
%            \hline
%         \end{tabular}
%      }
%   }   
%   \label{efficiency2}
%   \vspace{-0.3cm}
% \end{table}

\begin{table}
  \setlength{\belowcaptionskip}{.1cm}
  \setlength{\abovecaptionskip}{0.1cm}
  \caption{Inference speed of Baichuan-7B and EGG.}
  \centering
  \renewcommand\arraystretch{.9}
  \setlength{\tabcolsep}{8mm}{
  \scalebox{0.85}{
        \begin{tabular}{c|c}
           \hline    
           \toprule
              Methods&Inference Speed\\
           \midrule
           Baichuan-7B(Fact)       & 1.00 $\times$  \\
           
              \midrule
              $\textmd{EGG}_{free}$  & 0.71 $\times$  \\
              EGG  & 8.72 $\times$  \\
           \bottomrule
           \hline
        \end{tabular}
     }
  }   
  \label{efficiency2}
  % \vspace{-0.3cm}
\end{table}

Besides, we observe that Baichuan-7B without instruction tuning performs well on the legal task in the zero-shot setting, which indicates that Baichuan-7B has already possessed court view abilities through training on a large amount of data. However, its results are still worse than the fine-tuned model (Baichuan-7B(Fact)), which also shows the necessity of fine-tuning LLMs to the court view generation.
Although Baichuan-7B(Fact) achieves promising results on BLEU than EGG, it still performs worse on ROUGE and  Bert-S. Meanwhile,   EGG does not fine-tune LLMs in the generation phase, which  further illustrates the effectiveness of EGG.
Finally, we analyse the difference between EGG and $\textmd{EGG}_{free}$.
From Table~\ref{table1},  $\textmd{EGG}$ performs better than $\textmd{EGG}_{free}$, which indicates that it is more effective to combine case facts and events directly and explicitly on the data side than to introduce events implicitly into the model structure.
However, from the point of view of inference speed and computational resources occupied by the model, $\textmd{EGG}_{free}$ is faster and occupies fewer computational resources, yet achieves similar results to EGG.
This observation illustrates the effectiveness of $\textmd{EGG}_{free}$ which employs the contrastive learning constraint to incorporate event information into the learning of factual representations. In section \ref{efficiency}, we will further illustrate the efficiency of inference in $\textmd{{EGG}}_{{free}}$.

\subsection{Efficiency of Inference in $\textmd{\textbf{EGG}}_{\textbf{free}}$ (RQ2)}
\label{efficiency}

In this section, we present the results of our experiments comparing the inference speed of our proposed $\textmd{EGG}_{free}$ with other baselines. The hardware setup for the experiments consists of 12 cores of Intel(R) Xeon(R) Gold 5317 CPU and a single 40G NVIDIA A100 Tensor Core GPU. The findings are summarized in Table \ref{efficiency2}.
Our proposed $\textmd{EGG}{free}$ achieves an impressive decoding speed, approximately 12 times the speed achieved by EGG, which utilizes LLM for event extraction. It is worth noting that EGG has the slowest inference speeds. 
Besides, although the difference between the number of parameters in EGG and Baichuan-7B is not significant, since there are multiple events for a single case, EGG often needs to perform multiple event extractions, and thus is slower than Baichuan-7B.
This observation highlights that $\textmd{EGG}{free}$ strikes a balance between efficiency and effectiveness, making it well-suited for resource-constrained users.
The results demonstrate that $\textmd{EGG}_{free}$ offers a practical solution for legal event extraction, providing efficient performance while maintaining effectiveness. Its suitability for resource-constrained users makes it a valuable option in real-world applications.

% \subsubsection{Changes of gate values $g_{t}$}
% In this section, we make experiments to observe what the gate network actually learns.
% We show the changes of the gate values $g_{t}$ in the training set during training in Figure \ref{gate}.
% From the observation, at the beginning, the $g_{t}$ exceeds 0.5, which indicates the model is more biased to select the facts for generation.
% As training proceeds, this value begins to gradually decrease and stabilize at around 0.5, indicating that the model relies equally on the facts and the events when generating the court view. As events are extracted from the case facts, this further illustrates the importance of fine-grained event information in the court view~generation.
\subsection{Performance by the Length of Court Views (RQ3)}
In this section, we focus on investigating the generation performance of court views based on their length. We sample examples from the test set of CJO, where the real court views have lengths ranging from 50 to 120 tokens. We then predict and evaluate the generated court views by comparing them with the outputs of $\textmd{EGG}$ and C3VG with BART using ROUGE-L and BLEU-4 scores.
The findings, as illustrated in Figure \ref{rouge}, reveal that both $\textmd{EGG}$ and C3VG with BART experience a degradation in performance as the length of court views increases. However, we observe that our $\textmd{EGG}$ achieves the best performance when the court view length is between 60 and 70 tokens, with both ROUGE-L and BLEU-4 scores surpassing 90.
Furthermore, our method outperforms C3VG with BART across all court view lengths, indicating the effectiveness of incorporating fine-grained event information into court view generation. This suggests that by considering the specific event details in the generation process, our approach can produce more accurate and higher-quality court views compared to existing methods.

\subsection{Human Evaluation (RQ4)}

Table \ref{table1} highlights that both EGG and $\textmd{EGG}_{free}$ exhibit lower BLEU scores compared to Baichuan-7B(fact), prompting the need to investigate the performance of generated court views. To gain further insights, a human evaluation is conducted on the court views generated by EGG and Baichuan-7B(fact).
In this evaluation, a total of 100 examples are sampled, and three annotators with expertise in both computer science and law are asked to evaluate the generated court views based on two metrics: Usefulness and Fluency. Each metric is scored on a scale from 1 (lowest) to 5 (highest), with specific scoring standards provided in Table \ref{human}. The experimental results are presented in Table \ref{table_human}.
The results indicate that all models achieve promising scores in terms of Fluency, indicating that the generated court views are fluent and well-formed. Additionally, it is observed that EGG and $\textmd{EGG}_{free}$ outperform Baichuan-7B(fact) in terms of Usefulness. This finding further illustrates the effectiveness of incorporating fine-grained event information into court view generation. By considering the specific event details, our models generate court views that are deemed more useful by human evaluators.

% \subsection{T-SNE Visualization of Fact and  Event in $\textmd{\textbf{EGG}}_{\textbf{free}}$  (RQ5)}
% Besides, we 

\begin{table}
  \center
  \setlength{\belowcaptionskip}{.1cm}
  \setlength{\abovecaptionskip}{0.1cm}
  \caption{Detailed scoring standards for human annotators.}
% \vspace{0.2cm}
  \renewcommand\thetable{2}
  \setlength{\tabcolsep}{1.mm}{
    \scalebox{.8}{
  \renewcommand\arraystretch{1.}
  \begin{tabular}{p{20pt}p{180pt}p{70pt}}
  \toprule
      Score  & Usefulness   & Fluency \\
      \midrule
      1 & No Use. The generated texts are useless for answering questions. & Nonsense.\\
      2 & Almost useless. Almost all generated texts are useless.& Very unfluent.\\
      3 & Half of them are useful. About half of the generated texts are useful for answering questions.  & Partial fluent.\\
      4 & Highly useful. Most generated texts are useful to answer the questions. & Highly fluent.\\
      5 & Exactly. Generated texts are useful for me to get the correct answer. & Very fluent.\\

  \bottomrule
  \label{human}
\end{tabular}
    }}
    \vspace{-0.5cm}
\end{table}

\begin{table}
  \setlength{\belowcaptionskip}{.1cm}
  \setlength{\abovecaptionskip}{0.1cm}
  \caption{Human evaluation on generated texts.}
  \centering
  \renewcommand\arraystretch{.9}
  \setlength{\tabcolsep}{8mm}{
  \scalebox{0.85}{
        \begin{tabular}{c|cc}
           \hline    
           \toprule
              Methods&Usefulness&Fluency\\
           \midrule
           Baichuan-7B(fact)       & 3.76 &  4.43 \\
           
              \midrule
              $\textmd{EGG}_{free}$  & 3.98 &  4.63 \\
              EGG   & \textbf{4.09} &  \textbf{4.68} \\
           \bottomrule
           \hline
        \end{tabular}
     }
  }   
  \label{table_human}
\end{table}

\subsection{Case Study (RQ5)}
\begin{figure*}[htp]
  \centering 
  \setlength{\abovecaptionskip}{0.3cm}
  \setlength{\belowcaptionskip}{0cm}
  \includegraphics[width = 16.cm]{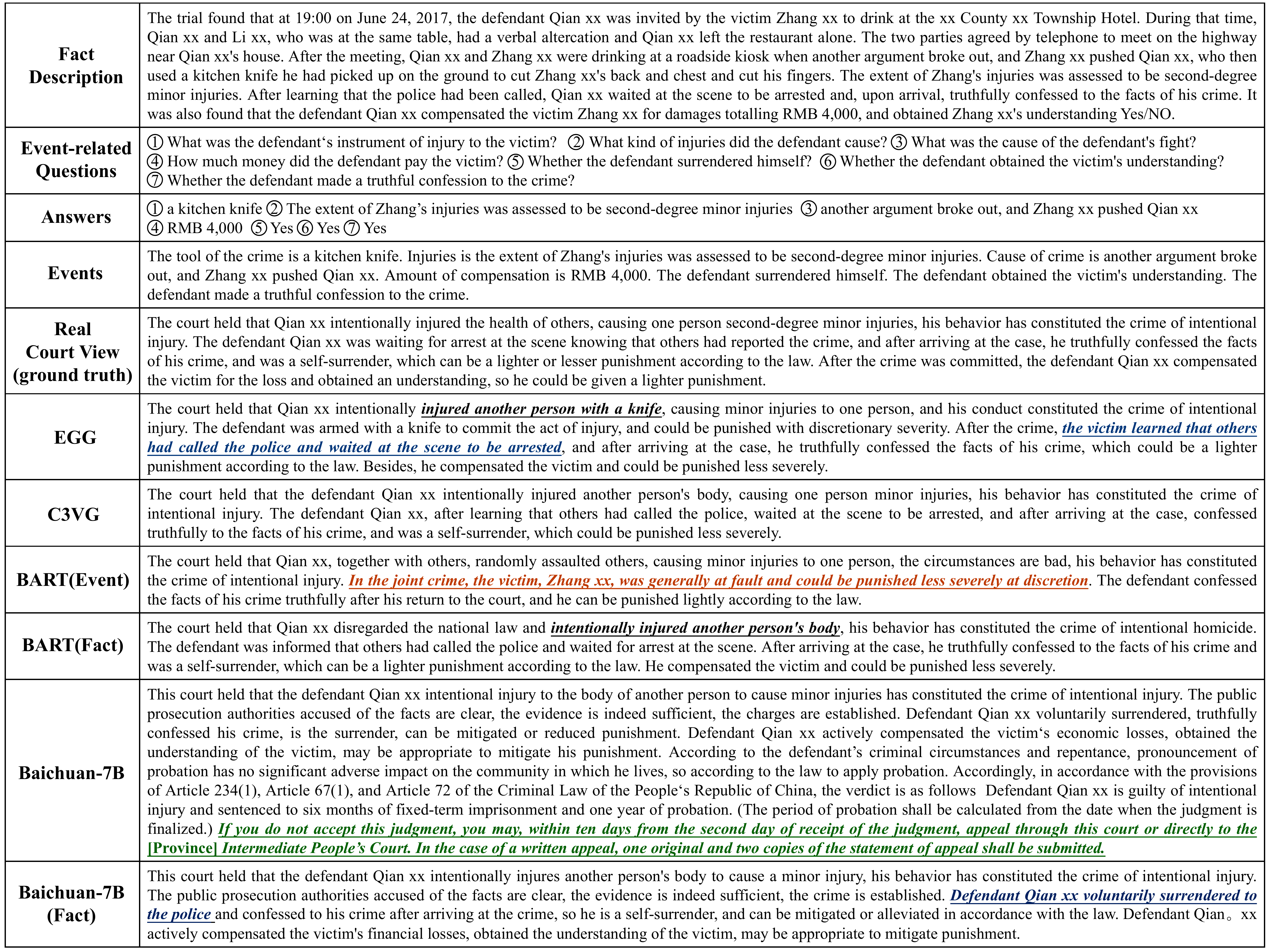}
 \caption{
  An example of extracted events and generated court views.
 }
 \label{casestudy}
 %\vspace{-0.3cm}
\end{figure*}

An example of extracted events and generated court views is shown in Figure \ref{casestudy}. 
Specifically, firstly, the type of case in this fact is \textit{intentional injury}. Then, \textbf{Event-related Questions} show all questions designed for the crime of intentional injury.
It is worth noting that the designed questions are the same for any fact which belongs to \textit{intentional injury}.
In the \textbf{Answers}, we present the answers extracted from the fact description according to the questions by the LLMs-based extractor. Afterward, we post-process the questions and answers to obtain the corresponding~\textbf{Events}.

Next, we present 7 court views generated by EGG and baselines:

$\bullet$
We can find although C3VG generates court views well, it fails to generate the court views about obtaining the victim's understanding.
Conversely, EGG can generate more plausible court views.
Besides, EGG also yields \textbf{\textit{\underline{injured another person with a knife}}} which has been described in fact description but not in the real court view. This observation indicates EGG can generate several key information, which is ignored by the real court view.

$\bullet$
Besides, compared to BART(Event), we find it generates several unfaithful court views (the underlined) which do not exist in the case fact, illustrating that it is unfeasible to generate court views based solely on events, and we need to combine the fact and event for the generation.
Then, although BART(Fact) has generated court views well, there are several omissions it generates compared to other methods. For example, EGG yields the injuries of the defendant is minor, however, BART(Fact) only generates the defendant has injuries~(i.e., \textbf{\textit{\underline{intentionally injured another person's body}}}).

$\bullet$
Moreover, we can get a fluent  court view by directly prompting Baichuan-7B (without fine-tuning). However, we tend to obtain some redundant information (e.g., ``\textit{If you do not accept this judgment, ... , directly to the [Province] Intermediate People's Court ...}''). This information has nothing to do with the court view, and may even involve some private information, such as \textit{[Province] Intermediate People's Court}, where we have a privacy treatment for [Province] to show this example. Meanwhile, Baichuan-7B(fact) is not as accurate as EGG in yielding information about surrender, where Baichuan-7B(fact) only describes the defendant's surrender,~while EGG also describes how the defendant surrendered (``\textit{the victim learned that others had called the police and waited at the scene to be arrested}'').
These observations demonstrate that incorporating fine-grained event information into the court view generation is effective.

\section{Discussion}
\textbf{Ethical Discussion.}
Court view generation has gained significant attention as a core task in legal intelligence.
Based on the experimental results, EGG demonstrates the ability to generate more accurate court views. Additionally, $\textmd{EGG}_{free}$ achieves a balance between modeling effectiveness and inference efficiency, making it suitable for users with limited computational resources.
However, it is important to note that our model does not replace the work of judges. Instead, our aim is to assist judges in organizing court views and alleviate their workload. The final court views must be determined and decided upon by the judges themselves \cite{wuetal2020de,yue2021circumstances}. Our work serves to provide judges with a tool to streamline the process of collating court views and reduce their workload stress.
% Moreover, our work also benefits laymen by providing them with quick insights into the case facts. Court views can be seen as summaries of the facts, enabling laymen to gain a better understanding of the case efficiently.
% In summary, our model acts as a supportive tool for judges, aiding them in the collation and organization of court views, while also providing laymen with easily accessible summaries of the case facts.

% With the development of artificial intelligence (AI), AI technology is applied in many legal  intelligent applications.
% Among them, court view generation has also attracted increasing attention as a fundamental task in legal intelligence.
% As observed from~the experimental results, our model can generate more accurate court views.
% Besides, $\textmd{EGG}_{free}$ also achieves a trade-off between modeling effectiveness and inference efficiency, and is well suited for resource-constrained users.
% However, it is worth noting that our model does not replace the work of the judges.
% On the contrary, our aim is to assist the judges in the collation of court views and to reduce the stress of workload of the judges. Final court views need to be determined by the judges~themselves \cite{wuetal2020de,yue2021circumstances}.
% Meanwhile, our work can provide laymen with quick insight into the case fact, where court views can be considered as the summary of facts.

\textbf{Limitations.}
When extracting events, we simply post-process the extracted answers and questions to obtain the corresponding events.  However, when dealing with complex relationships among events, such as causality, a more advanced approach is needed. One possible solution is to construct an event graph that represents the relationships among events.
  An event graph is a graphical representation where events are nodes, and the relationships between events are represented by edges. By incorporating this event graph into the court view generation process, the model can better capture and understand the complex relationships among events.
  We will leave it as the future work.

\section{Conclusion}
In this paper, we proposed an Event Grounded Generation (EGG) method for criminal court view generation with cooperative (Large) Language Models, cascading the event extractor and the court view generator. 
To be specific, EGG first employed a trained LLMs-based legal  event extractor to select several events in the case fact without massive annotated events.
Then, in the court view generator, we incorporated these events into the court view generation by merging the case fact and event as the new input.
Besides, to alleviate the computational burden in EGG during inference that employs LLMs, we further proposed a LLMs-free EGG method based on the contrastive constraint. This enhancement enables court view generation without requiring event information during the inference phase.
Experimental results on a real-world dataset clearly demonstrated the effectiveness of our proposed method.

\textbf{Acknowledgements.} This research was supported by grants from the National Natural Science Foundation of China (Grants No. 62337001, 623B1020) and the Fundamental Research Funds for the Central Universities.
\bibliographystyle{ACM-Reference-Format}
\bibliography{ref}

\end{document}